\documentclass[final]{cvpr}
\usepackage{times}
\usepackage{epsfig}
\usepackage{graphicx}
\usepackage{amsmath}
\usepackage{amssymb}
\usepackage{stfloats}
\usepackage{array}
\usepackage{multirow}
\usepackage{booktabs}
\usepackage{color}

\usepackage[pagebackref=true,breaklinks=true,colorlinks,bookmarks=false]{hyperref}

\begin{document}

\title{Think about boundary: Fusing multi-level boundary information for landmark heatmap regression}

\author{Jinheng Xie\quad Jun Wan\quad Linlin Shen \quad Zhihui Lai\\
 Computer Vision Institue, School of Computer Science \& Software Engineering, Shenzhen University \\
{\tt\small sierkinhane@163.com, junwan2014@whu.edu.cn, llshen@szu.edu.com, lai\_zhi\_hui@163.com }}

\maketitle

\begin{abstract}
Although current face alignment algorithms have obtained pretty good performances at predicting the location of facial landmarks, huge challenges remain for faces with severe occlusion and large pose variations, etc. On the contrary, semantic location of facial boundary is more likely to be reserved and estimated on these scenes. Therefore, we study a two-stage but end-to-end approach for exploring the relationship between the facial boundary and landmarks to get boundary-aware landmark predictions, which consists of two modules: the self-calibrated boundary estimation (SCBE) module and the boundary-aware landmark transform (BALT) module. In the SCBE module, we modify the stem layers and employ intermediate supervision to help generate high-quality facial boundary heatmaps. Boundary-aware features inherited from the SCBE module are integrated into the BALT module in a multi-scale fusion framework to better model the transformation from boundary to landmark heatmap. Experimental results conducted on the challenging benchmark datasets demonstrate that our approach outperforms state-of-the-art methods in the literature. The code and models will be available at \url{https://github.com/CVI-SZU/TAB}
\end{abstract}

%%%%%%%%% BODY TEXT
\section{Introduction}
\begin{figure}[ht]
	\begin{center}
		%\fbox{\rule{0pt}{2in} \rule{.9\linewidth}{0pt}}
		\includegraphics[width=\columnwidth]{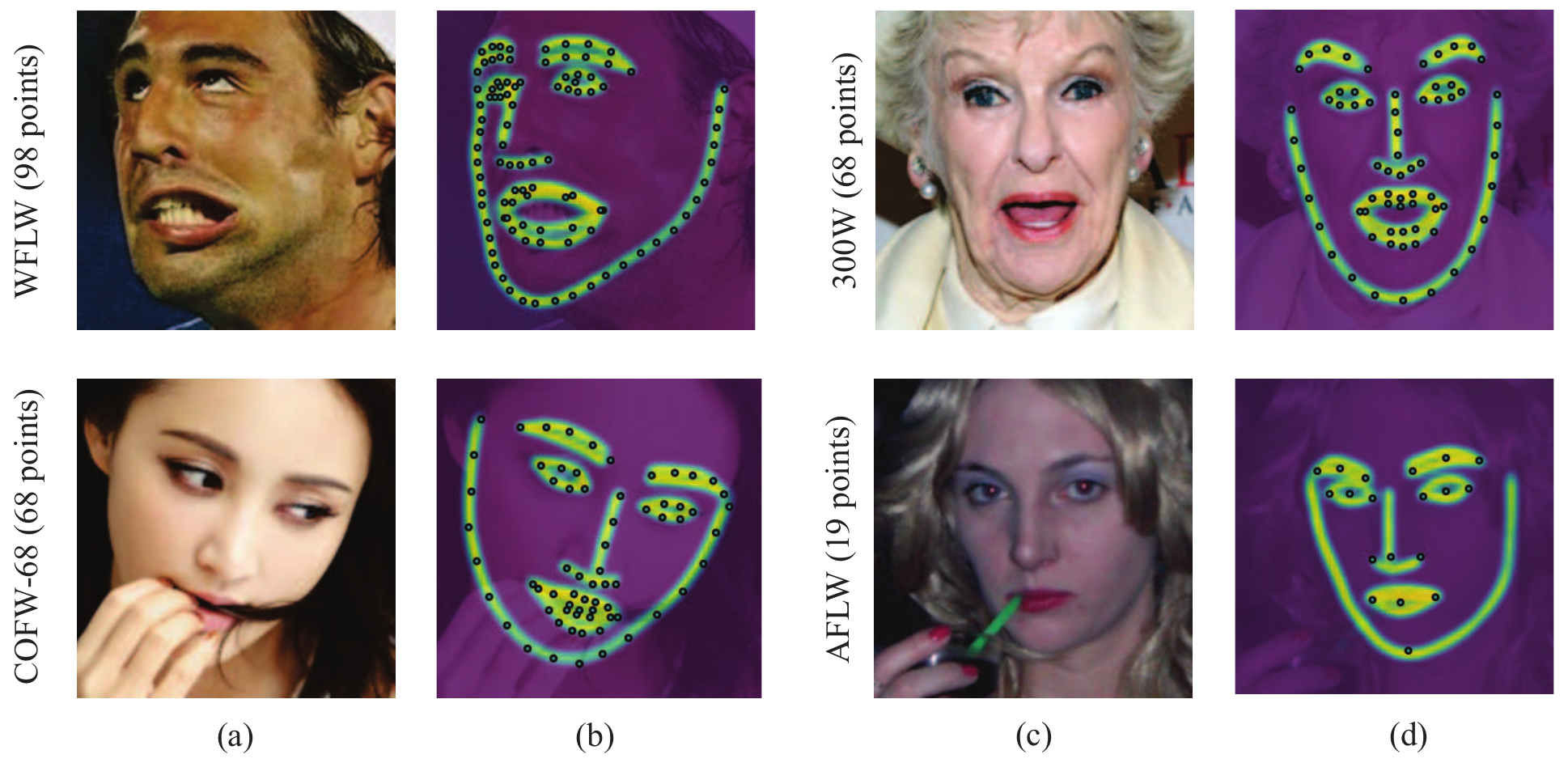}
	\end{center}
%	\vspace*{-0.1in}
	\caption{Examples of facial landmark annotations and facial boundaries across different datasets. The WFLW, COFW-68, 300W, and AFLW datasets were labeled with 98 landmarks, 68 landmarks, 68 landmarks, and 19 landmarks, respectively. For all datasets, we use 15 boundaries to represent their boundary information.}
	\label{fig:facial_annotation}
%	\vspace*{-0.1in}
\end{figure}

The task of face alignment, i.e. facial landmark localization, is to detect a pre-defined keypoints of 2D face images, which is an essential pre-processing procedure for many face applications such as face recognition~\cite{Schroff_2015_CVPR}~\cite{Deng_2019_CVPR}, face reconstruction~\cite{tu20203d}~\cite{feng2018joint}, facial expression analysis and face edition. Generally, there is a semantic geometric structure meaning of facial landmarks, such as cheek, lips, and eyebrows, etc., as shown in Fig.~\ref{fig:facial_annotation}. These geometric structure information of faces is more likely to be reserved and estimated on unconstrained and complicated scenes, such as faces with occlusion, large pose, and make-up, etc. 
\begin{figure}
	\begin{center}
		%\fbox{\rule{0pt}{2in} \rule{0000000000.9\linewidth}{0pt}}
		\includegraphics[width=\columnwidth]{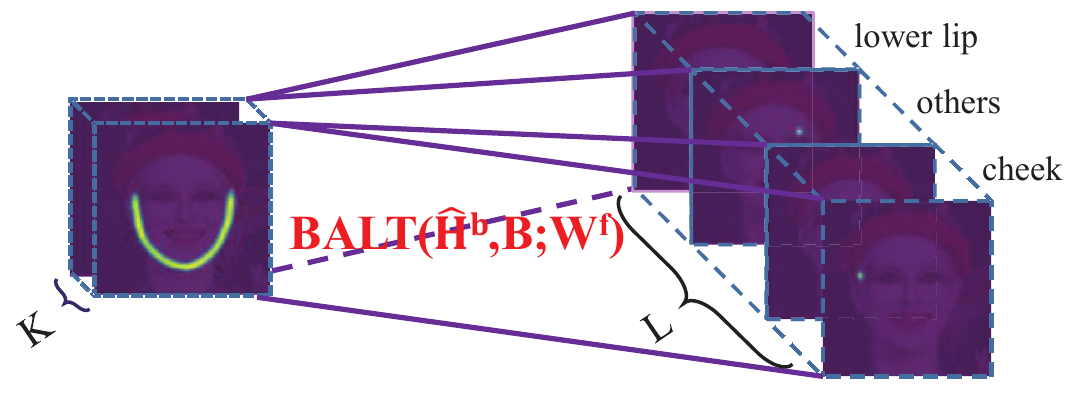}
	\end{center}
%	\vspace*{-0.2in}
	\caption{Boundary-aware landmark transform. The location of facial landmarks rely on the facial boundary, it is apparently that the landmark heatmaps can be obtained by transforming these facial boundary heatmaps. Therefore, the BALT module was proposed to estimate this mapping relationship to enhance the shape constraint of facial boundary on facial landmarks.}
	\label{fig:HWT}
%	\vspace*{-0.1in}
\end{figure}

The definition of the facial boundary is proposed by Wu, et al.~\cite{wayne2018lab}, they design a two-stage method to regress the coordinates of facial landmarks. The approach  proposes a boundary heatmap estimator for extracting the boundary information of faces and a boundary-aware landmark regressor for regressing the coordinates of facial landmarks. However, we argue that using input heatmap fusion to highlight the boundary area of image and feature map fusion to directly regress the coordinates of facial landmarks does not make full use of the facial boundary information. There are four reasons for this claim: (1) In the landmark regressor, using input image fusion to highlight facial boundary will increase the channels of input, which significantly increase the computation and complexity of shallow convolutional layers. And the rich boundary-aware features extracted by the boundary estimator are not fully explored. (2) They use a network to directly regress the coordinates of faces, which did not take into
account the advantage of the spatial constraints bettween facial landmarks. To learn a transformation bettween boundary and landmark heatmaps(matrix-matrix mapping) to get the final landmark predictions seems more reasonable (illustrated by Fig.~\ref{fig:HWT}). (3) At the end of the landmark regressor, the downsampled heatmaps will impair the shape constraint on facial landmark predictions, due to the lost boundary information. To make full use of the rich faical boundary information, we investigate a two-stage but end-to-end method for exploring the relationship between facial boundary and landmarks to get boundary-aware landmark heatmap predictions, in which boundary-aware features can be used to accelerate the convergence of training and enhance the shape constraint of facial boundary on facial landmark heatmap predictions. 

Our approach mainly consists of two modules, the self-calibrated boundary estimation (SCBE) and the boundary-aware landmark transform (BALT). Based on the Stacked Hourglass Network~\cite{newell2016stacked}, SCBE is mainly used to estimate the boundary information of faces. To improve the performance of SCBE module on faces with severe occlusion and large
pose, etc., we replace the stem layers of original Stacked Hourglass Network with VGG~\cite{simonyan2014very}/ResNet~\cite{he2016deep}, and initialize them with parameters pre-trained on ImageNet~\cite{deng2009imagenet} to accelerate the convergence of training. Moreover, SCBE module can self-calibrate the boundary heatmap predictions by introducing intermediate supervision. The BALT module is an Encode-Decode Network based on UNet~\cite{ronneberger2015u}, which can inherit the multi-level boundary information from SCBE module to obtain more accurate landmark heatmaps in a transformation manner. Compared to mapping heatmaps to coordinates, we believe that neural network is more easily to learn the mapping between boundary heatmaps and landamark heatmaps, due to their same spatial representation. In addition, we integrate a multi-scale feature fusion module into the encoding process to fully use the boundary-aware features extracted by the SCBE module. A multi-level shape constraint enhancement is also proposed to enhance the shape constraint of facial boundary on landmarks to get boundary-aware landmark heatmap predictions. Finally, the mean squared error is adopted to combine the two modules to obtain an end-to-end process. The main contributions of this paper can be summarized as below:
\begin{figure}
	\begin{center}
		%\fbox{\rule{0pt}{2in} \rule{0000000000.9\linewidth}{0pt}}
		\includegraphics[width=\columnwidth]{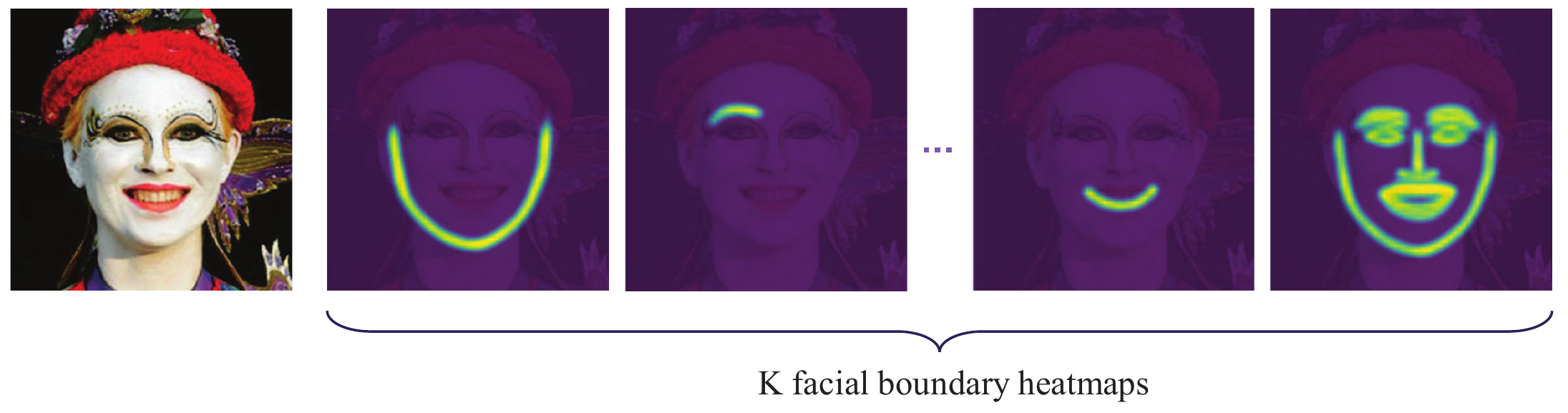}
	\end{center}
%		\vspace*{-0.15in}
	\caption{Facial boundary heatmaps. 15 boundaries are used to represent the boundary of the face, and the 16th  heatmap is a union of all boundaries.}
	\label{fig:16_boundaries}
%	\vspace*{-0.15in}
\end{figure}
\begin{enumerate}
	\item[-] We study the effect of stem layers and intermediate supervision for facial boundary estimation to improve the SCBE module's robustness on unconstrained faces and boost the performance of the BALT module. 
	\item[-] Instead of directly using boundary heatmaps to regress the coordinates of facial landmarks, we propose to map boundary heatmaps to landmark heatmaps to enhance the efficiency of transformation, due to their same spatial representation.
	\item[-] We propose  multi-scale features fusion to make full use of the boundary-aware features extracted by the SCBE module, which accelerate the convergence of training and improve the efficiency of transformation from boundary to landmark.  We also introduce the multi-level shape constraint enhancement to enhance the shape constraint of facial boundary on landmarks to get boundary-aware heatmap predictions.
	\item[-] We evaluate our method on the widely used face alignment benchmarks and achieve state-of-the-art performance. Especially on the WFLW dataset, our method achieves significant improvement over state-of-the-art methods.
\end{enumerate}

%------------------------------------------------------------------------
\section{Related works}

 Active shape model (ASM)~\cite{milborrow2008locating} and active appearance model (AAM)~\cite{cootes2001active, saragih2007nonlinear} are milestones of traditional face alignment methods. With rapid development of the Convolutional Neural Networks (CNNs)~\cite{lecun1998gradient}, more and more CNN-based face alignment methods were proposed. These methods generally fall into two categories: numerical regression-based  and heatmap regression-based methods.

\textbf{Numerical regression-based methods} leverage the feed-forward framework to get a fast prediction. Its input is a 2D image and output is numerical values. These methods aim to learn a mapping from image to the coordinates of landmarks. Sun et al.~\cite{sun2013deep} use the convolutional neural network for the first time to detect the landmarks of human faces and propose a three-level cascaded convolutional neural network to obtain the precise position of facial landmarks. Zhang et al.~\cite{7553523} utilize three successive networks to get the coordinates of the bounding box and five facial landmarks. Guo et al.~\cite{guo2019pfld} customize an end-to-end network to estimate rotation of head for regularizing landmark localization to predict landmark coordinates, which achieves a fast speed with reasonable performance. Feng et al.~\cite{feng2018wing} design a novel piece-wise loss that can pay more attention on small and medium errors to accelerate the convergence of training and obtain a good performance. Wu et al.~\cite{wayne2018lab} propose boundary heatmaps to help the neural network to regress the coordinates of facial landmarks. Wan et al.~\cite{wan2020robust} aim to improve the performance on occluded faces via a face de-occlusion module and a deep regression module. Nevertheless, these methods are generally fast, but not as good as the heatmap regression-based methods.

\textbf{Heatmap regression-based methods} show a better performance than numerical methods. Their final output are heatmaps, which have the same representation with an image in 2D pixels. It is more likely to utlize the spatial location constraint between pixels to learn a mapping from an image to a heatmap. Kowalski et al.~\cite{kowalski2017deep} design a cascaded network and use the heatmap for face alignment for the first time. Dapogny et al.~\cite{dapogny2019decafa} design a cascaded-UNet to keep the full spatial resolution to regress landmark heatmaps with intermediate supervision. Wang et al.~\cite{wang2019adaptive} devise a novel loss function to adapt its shape to different types of pixels in the ground truth heatmap, which obtains better regression. Heatmap based methods are also widely used in other vision tasks, such as human pose estimation~\cite{cao2018openpose}\cite{fang2017rmpe}\cite{sun2019deep} and object detection~\cite{law2018cornernet}, etc.

\textbf{With extra boundary information.} Wu et al.~\cite{wayne2018lab} introduce boundary heatmaps to assist landmark coordinates regression. It firstly estimates facial boundary, then input image fusion and features fusion are performed at the beginning of the landamrk regressor to regress the coordinates of facial landmarks. While original face image is processed twice in both boundary heatmap and landmark regression, the shape constraint of boundary on landmark is actually weak, due to the downsampling process. The more advanced heatmap regression is not explored as well. Wang et al.~\cite{wang2019adaptive} simply include the boundary heatmap as the output of network, which plays as a weak supervision to integrate the boundary information. To fully integrate the useful boundary information into the most recent heatmap based landmark regression, we propose a two-stage end-to-end method to explore the relationship between facial boundary and landmark heatmaps. The boundary-aware features learned in the boundary heatmap estimation network are fully explored to enhance the shape constraint of facial boundary on landmark heatmap predictions. 

%-------------------------------------------------------------------------
\section{Methodology}

\begin{figure*}
	\begin{center}
		%\fbox{\rule{0pt}{2in} \rule{.9\linewidth}{0pt}}
		\includegraphics[width=\linewidth]{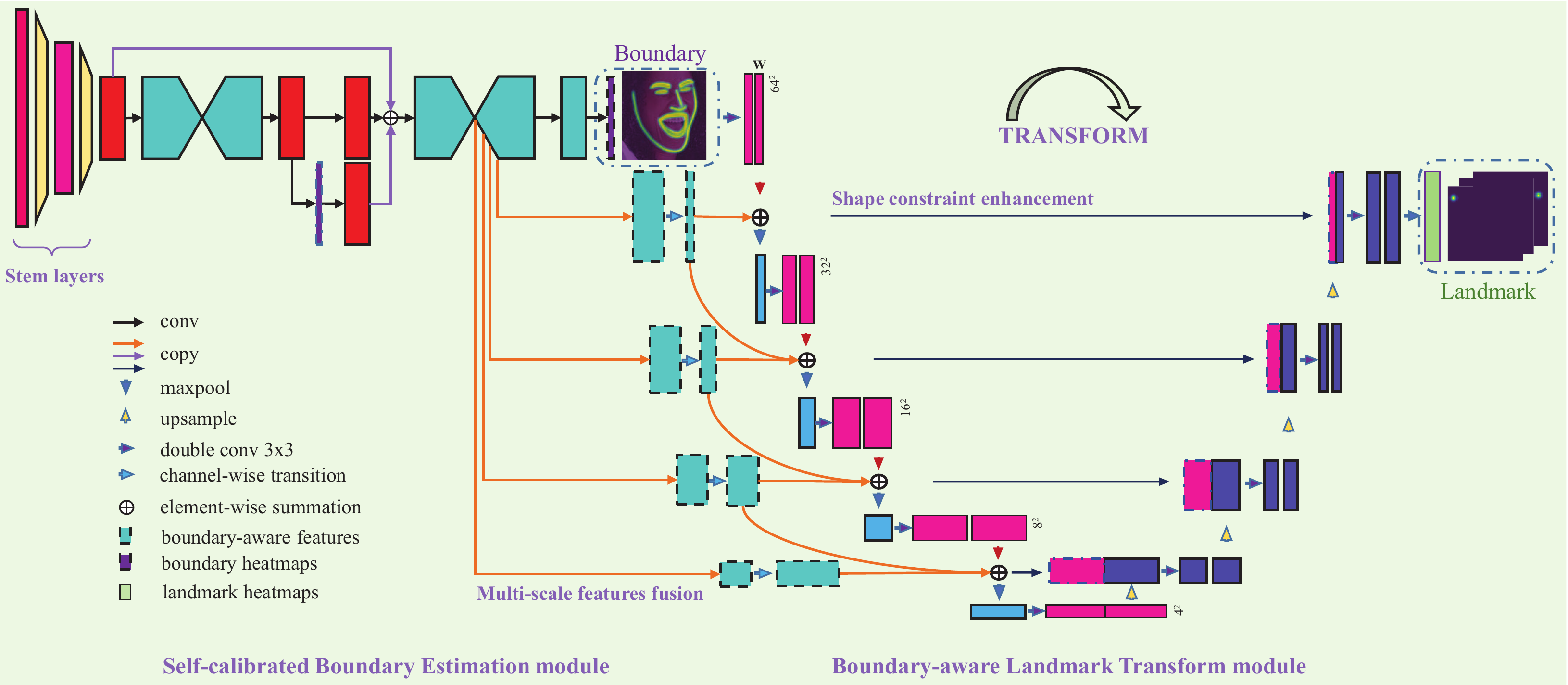}
	\end{center}
	\vspace*{-0.05in}
	\caption{The overall architecture of our  model. The SCBE module is used to estimate the facial boundary information of faces. The BALT module is used to learn how to transform the boundary to landmark to get the boundary-sensitive landmark predictions with multi-scale features fusion. The multi-scale features are inherited from the last hourglass of the SCBE module, which encodes a lot of boundary-aware features.}
	\label{fig:BATNet}
	\vspace*{-0.1in}
\end{figure*}

The proposed method contains two modules: the SCBE module and the BALT module. The SCBE module aims to estimate the facial boundary heatmaps of images. Then, the BALT module takes these boundaries as input and learns a matrix-matrix mapping to transform facial boundary heatmaps to landmark heatmaps with multi-level boundary information fusion. An overview of the proposed method is shown in Fig ~\ref{fig:BATNet}. In the following, we provide more details about each module.

%-------------------------------------------------------------------------
\subsection{Self-calibrated boundary estimation module}

Wu et al.~\cite{wayne2018lab} use an N-Stacked Hourglass Network(N=4 or 8) and a boundary effectiveness discriminator to train a facial boundary estimation model. To simplify the training process and slim the model, we use a 2-Stacked Hourglass Network and emphasize the mutual effect of stem layers and intermediate supervision, which can refine the final boundary heatmap predictions in a self-calibration manner. 

As shown in Fig.~\ref{fig:16_boundaries}, we use $K$ boundary heatmaps to represent the faicial boundary information of an image, which include $K-1$ facial boundaries such as cheek, lips, and eyebrows, etc., and the $K$th heatmap, i.e. a union of all boundaries. The SCBE module can be expressed as follows:
%and only use WFLW dataset~\cite{wayne2018lab} to train our SCBE module
\begin{equation}
\hat{\mathbf{H}}^{b} = \mathcal{SCBE}(\mathbf{I}; \mathbf{W}^{s*}, \mathbf{W}^h)
\end{equation}
Our SCBE module aims to estimate the facial boundary heatmaps $\hat{\mathbf{H}}^{b}$ from an image $\mathbf{I}$, $ \mathbf{W}^{s*}$ denotes the parameters pre-trained on ImageNet dataset and $\mathbf{W}^h$ denotes the learnable paramters of the N-Stacked Hourglass.

We name the shallow convolutional layers with output from $256\times 256$ to $64\times 64$ as stem layers (illustrated on the left-most of Fig.~\ref{fig:BATNet}), which can extract more detailed information such as edges and further help to  learn more discriminative representations on subsequent convolutional layers. To increase the robustness of stem layers and extract more rich and accurate semantic information, we try to use ResNet and VGG to replace the stem layers of the Hourglass Network designed by~\cite{bulat2017far}, and initialize them with parameters $\mathbf{W^{s*}}$ pre-trained on ImageNet dataset. The experimental results shown in Table~\ref{table:stemlayers} indicate that, compared with Hourglass and ResNet,  VGG achieves lower loss and better performance. See section~\ref{subsection:as} for more details.  

While SCBE module only stacks two hourglasses, we also introduce two effective approaches to improve the quality of predicted heatmaps (illustrated by purple solid line on the left-most of Fig.~\ref{fig:BATNet}): (1) we extract the low-level features from the former layer and reuse it in the next hourglass to enhance the feature representation. (2) we add a boundary heatmap predictions between the two-stacked hourglass to  self-calibrate the final boundary heatmap predictions via intermediate supervision.

%-------------------------------------------------------------------------
\subsection{Boundary-aware landmark transform module}

While heatmaps are a set of discrete 2D representations of landmarks, boundary heatmap is a continuous representation of these landmarks and is thus more robust to large pose variations and occlusions. It seems to be more resonable to learn the transformation between the two heatmaps, than the transformation between boundary heatmap and landmark coordinates. The transformation process can be expressed as follows:
\begin{equation}
\hat{\mathbf{H}}^{l} = \mathcal{BALT}(\hat{\mathbf{H}}^b, \mathbf{B}; \mathbf{W}^{f})
\end{equation}
%And its output -- boundary heatmaps and rich boundary-aware semantic features learned by hourglass are expected to be exploited.
$\hat{\mathbf{H}}^{l}$  denotes the predicted facial landmark heatmaps, $\hat{\mathbf{H}}^{b}$ denotes the predicted facial boundary heatmaps, $\mathbf{B}$ denotes the boundary-aware features extracted by the SCBE module,  and $\mathbf{W}^{f}$ denotes the learnable parameters of the BALT module. 

In above section, due to the stem layers and intermediate supervision, our SCBE module obtains refined and accurate boundary heatmaps and learns rich semantic boundary-aware features. To learn a transformation between boundary heatmap and landmark heatmap, we keep the resolution of predicted boundary heatmaps and use it as the input of the BALT module. Then, we fuse the rich and multi-scale boundary-aware features into the encoding and decoding process to enhance the feature representation and the facial boundary constraint on landmark heatmap predictions.

\begin{figure}
	\begin{center}
		%\fbox{\rule{0pt}{2in} \rule{0000000000.9\linewidth}{0pt}}
		\includegraphics[width=\columnwidth]{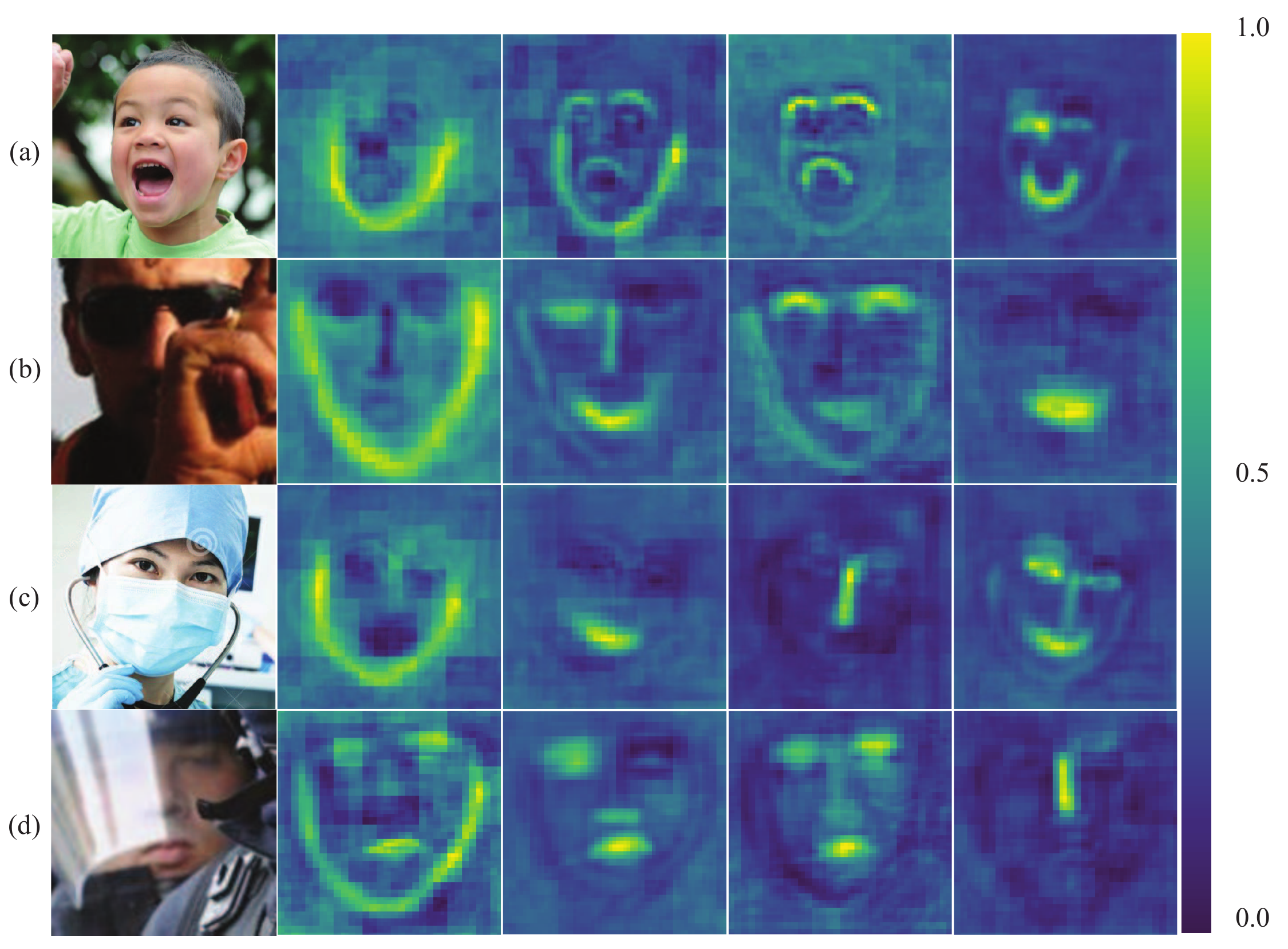}
	\end{center}
	%	\vspace*{-0.2in}
	\caption{Boundary-aware feature maps. We visualize the 64$\times$64 output of the last hourglass block of the SCBE module, which presents a good perception of the boundaries of  cheek, nose bridge, and lips, for faces with heavy occlusions.}
	\label{fig:boundary_aware_features}
		\vspace*{-0.15in}
\end{figure}

\textbf{How to choose boundary-aware features?} While the feature maps of the last block of SCBE contains high level boundary-aware features, multi-scale features extracted from the last hourglass block (the last light green bock in Fig.~\ref{fig:BATNet}) could be a good choice to provide more rich information. We visualize the feature maps (64$\times$64) of the last block of SCBE for four example faces (as shown in Fig.~\ref{fig:boundary_aware_features}). One can observe from the figure that they have good perception of boundary in cheek, lips and nose bridge, even when the faces are heavily occluded by hand and masks. We adaptively fuse these multi-level features  $\{\mathbf{B}_1,\mathbf{B}_2,\cdots,\mathbf{B}_s\}$ and inject them to different levels of the encoder. 

\textbf{How to use these boundary-aware features?}
Let $\{\mathbf{F}_1^e,\mathbf{F}_2^e,\cdots,\mathbf{F}_s^e\}$ represent the outputs of the BALT module in the encoding process, whose resolutions are the same to the s boundary-aware feature maps $\mathbf{B}$. Each feature map is passed into a channel-transition layer $\mathcal{T}$, then processed with element-wise summation $\oplus$, and finally passed into the maxpooling layer $\mathcal{MP}$ and the double convolutional layers $\mathcal{DC}$:
\begin{equation}
	\mathbf{F}_{i+1}^e=\mathcal{DC}(\mathcal{MP} (\mathbf{F}_i^e\oplus \mathcal{T}(\mathbf{B}_i) ))
\end{equation}  
It is the $1\times$ fusion process. For $t \times$ fusion, we add shortcuts between the neighboured convolutinal layers with different resolutions. The $i$th fusion process can be formulated as follows::
\begin{equation}\mathbf{F}_{i+1}^e=\left\{\begin{array}{ll}
	\mathcal{DC}(\mathcal{MP} (\mathbf{F}_i^e \oplus \mathbf{B}_i^{'})), & (i=1) \\
	\mathcal{DC}(\mathcal{MP}(\mathbf{F}_i^e \oplus\mathbf{B}_i^{'}\oplus \sum_{j=m}^{i-1}\mathbf{B_j^{''}} )), & (i>1) 
\end{array}\right.\end{equation} 
where $\mathbf{B}_i^{'}$ represents $\mathcal{T}(\mathbf{B}_i)$, $\mathbf{B}_j^{''}$ represents
 $\mathcal{MP}(\mathcal{T}(\mathbf{B}_j))$, and $m=\mathbf{Max}(1,i-t+1)$. In the decoding layers, we introduce a shape constraint enhancement module. Let $\{\mathbf{F_1}^d,\mathbf{F}_2^d,\cdots,\mathbf{F}_s^d\}$ represent the outputs of the BALT module in the decoding process, whose resolution are the same to the inverted order of s feature maps $\mathbf{B}$. We illustrate the shape constraint enhancement as: 
\begin{equation}
\mathbf{F}_{i+1}^d = \mathcal{BU}(\mathcal{DC}(\mathbf{F}_{s-i+1}^{e'}\otimes \mathbf{F}_i^d))
\end{equation}
where $\mathcal{BU}$ denotes the bilinear upsample,  $\mathcal{DC}$ denotes the double convolutional layers, $\mathbf{F}_{s-i+1}^{e'}$ denotes the fusion output before maxpooling layer and double convolutional layers, and $\otimes$ denotes the channel-wise concatenation.
\begin{figure*}
	\begin{center}
		\includegraphics[width=\textwidth]{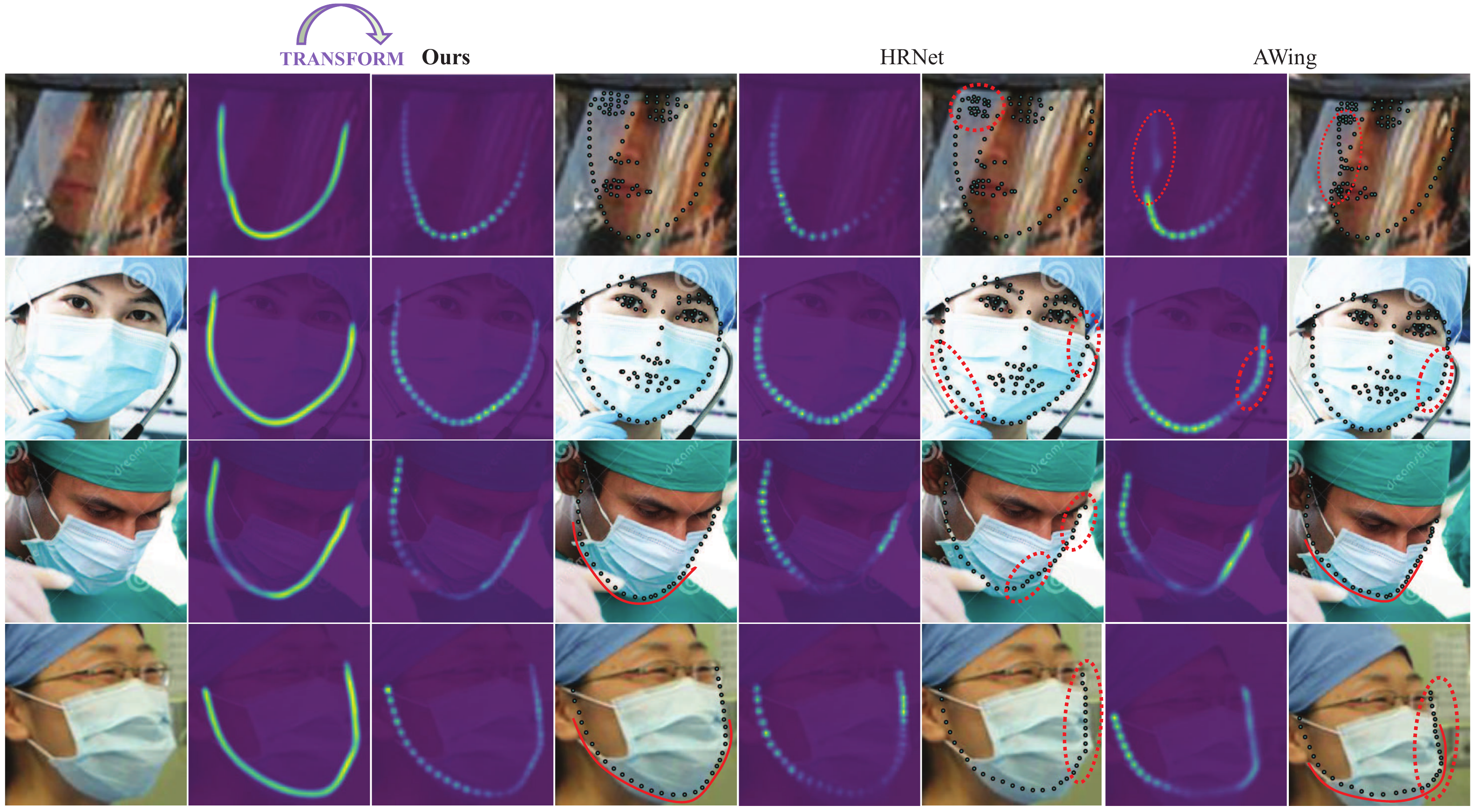}
	\end{center}
	%	\vspace*{-0.1in}
	
	\caption{Examples of facial boundary adaptable landmark predictions for different methods. Compared to HRNet and AWing, our method achieved boundary-sensitive, smoother, and more uniformed landmark predictions.}
	\label{fig:transition_example}
	%	\vspace*{-0.2in}
\end{figure*}

\begin{table*}
	%	\vspace*{-0.1in}
	\centering
	\caption{Evaluation on the WFLW dataset.}
	\resizebox{\linewidth}{!}{%
		\begin{tabular}{c|c|c|c|c|c|c|c|c}
			\hline
			Metric                            & Method & Testset & \begin{tabular}[c]{@{}c@{}}Pose\\ Subset\end{tabular} & \begin{tabular}[c]{@{}c@{}}Expression\\ Subset\end{tabular} & \begin{tabular}[c]{@{}c@{}}Illumination\\ Subset\end{tabular} & \begin{tabular}[c]{@{}c@{}}Make-up\\ Subset\end{tabular} & \begin{tabular}[c]{@{}c@{}}Occlusion\\ Subset\end{tabular} & \begin{tabular}[c]{@{}c@{}}Blur\\ Subset\end{tabular} \\ \hline
			\multirow{9}{*}{NME(\%)($\downarrow$)}   & ESR\textsubscript{CVPR 14}~\cite{cao2014face}   & 11.13  & 25.88  & 11.47  & 10.49  & 11.05  & 13.75  & 12.20  \\ 
			& SDM\textsubscript{CVPR 13}~\cite{xiong2013supervised}   & 10.29  & 24.10  & 11.45  & 9.32   & 9.38   & 13.03  & 11.28  \\
			& CFSS\textsubscript{CVPR 15}~\cite{zhu2015face}  & 9.07   & 21.36  & 10.09  & 8.30   & 8.74   & 11.76  & 9.96   \\
			& DVLN\textsubscript{CVPR 17}~\cite{wu2017leveraging}  & 6.08   & 11.54  & 6.78   & 5.73   & 5.98   & 7.33   & 6.88   \\
			& LAB\textsubscript{CVPR 18}~\cite{wayne2018lab}   & 5.27   & 10.24  & 5.51   & 5.23   & 5.15   & 6.79   & 6.32   \\
			& Wing\textsubscript{CVPR 18}~\cite{feng2018wing}  & 5.11   & 8.75   & 5.36   & 4.93   & 5.41   & 6.37   & 5.81   \\
			& HRNet\textsubscript{19'}~\cite{sun2019high}  & 4.60   & 7.86   & 4.78   & 4.57   & 4.26   & 5.42   & 5.36   \\
			& \text{AWing}\textsubscript{ICCV 19}~\cite{wang2019adaptive} & \text{4.36} & \text{7.38} & \text{4.58} & \text{4.32} & \text{4.27} & \text{5.19} & \text{4.96} \\ 
			& LUVLi\textsubscript{CVPR 20}~\cite{kumar2020luvli}  & 4.37   & 7.56   & 4.77   & 4.30   & 4.33  & 5.29   & 4.94   \\
			&\textbf{TAB(Ours)} & \textbf{3.94} & \textbf{6.70} & \textbf{4.11} & \textbf{3.84} & \textbf{3.85} & \textbf{4.56} & \textbf{4.45} \\
			\hline
			\multirow{9}{*}{FR\textsubscript{10\%}(\%)($\downarrow$)} & ESR\textsubscript{CVPR 14}~\cite{cao2014face}   & 35.24  & 90.18  & 42.04  & 30.80  & 38.84  & 47.28  & 41.40  \\
			& SDM\textsubscript{CVPR 13}~\cite{xiong2013supervised}   & 29.40  & 84.36  & 33.44  & 26.22  & 27.67  & 41.85  & 35.32  \\
			& CFSS\textsubscript{CVPR 15}~\cite{zhu2015face}  & 20.56  & 66.26  & 23.25  & 17.34  & 21.84  & 32.88  & 23.67  \\
			& DVLN\textsubscript{CVPR 17}~\cite{wu2017leveraging}  & 10.84  & 46.93  & 11.15  & 7.31   & 11.65  & 16.30  & 13.71  \\
			& LAB\textsubscript{CVPR 18}~\cite{wayne2018lab}   & 7.56   & 28.83  & 6.37   & 6.73   & 7.77   & 13.72  & 10.74  \\
			& Wing\textsubscript{CVPR 18}~\cite{feng2018wing}  & 6.00   & 22.70  & 4.78   & 4.30   & 7.77   & 12.50  & 7.76   \\
			& \text{AWing}\textsubscript{ICCV 19}~\cite{wang2019adaptive} & \text{2.84} & \text{13.50} & \text{2.23} & \text{2.58} & \text{2.91} & \text{5.98} & \text{3.75}   \\
			& LUVLi\textsubscript{CVPR 20}~\cite{kumar2020luvli}  & 3.12   & 15.95   & 3.18   & 2.15   & 3.40  & 6.39   & 3.23   \\
			&\textbf{TAB(Ours)} & \textbf{1.96} & \textbf{8.59} & \textbf{1.59} & \textbf{1.58} & \textbf{1.94} & \textbf{3.94} & \textbf{2.20} \\ \hline
			\multirow{9}{*}{AUC\textsubscript{10\%}($\uparrow$)}              & ESR\textsubscript{CVPR 14}~\cite{cao2014face}   & 0.2774 & 0.0177 & 0.1981 & 0.2953 & 0.2485 & 0.1946 & 0.2204 \\
			& SDM\textsubscript{CVPR 13}~\cite{xiong2013supervised}   & 0.3002 & 0.0226 & 0.2293 & 0.3237 & 0.3125 & 0.2060 & 0.2398 \\
			& CFSS\textsubscript{CVPR 15}~\cite{zhu2015face}  & 0.3659 & 0.0632 & 0.3157 & 0.3854 & 0.3691 & 0.2688 & 0.3037 \\
			& DVLN\textsubscript{CVPR 17}~\cite{wu2017leveraging}  & 0.4551 & 0.1474 & 0.3889 & 0.4743 & 0.4494 & 0.3794 & 0.3973 \\
			& LAB\textsubscript{CVPR 18}~\cite{wayne2018lab}   & 0.5323 & 0.2345 & 0.4951 & 0.5433 & 0.5394 & 0.4490 & 0.4630 \\
			& Wing\textsubscript{CVPR 18}~\cite{feng2018wing}  & 0.5504 & 0.3100 & 0.4959 & 0.5408 & 0.5582 & 0.4885 & 0.4918 \\
			& \text{AWing}\textsubscript{ICCV 19}~\cite{wang2019adaptive} & \text{0.5719} & \text{0.3120}   & \text{0.5149} & \text{0.5777} & \text{0.5715} & \text{0.5022} & \text{0.5120} \\ 
			& LUVLi\textsubscript{CVPR 20}~\cite{kumar2020luvli}  & 0.577   & 0.310   & 0.549   & 0.584   & 0.588  & 0.505   & 0.525   \\
			&\textbf{TAB(Ours)} & \textbf{0.6112} & \textbf{0.3577} & \textbf{0.5979} & \textbf{0.6203} & \textbf{0.6205} & \textbf{0.5552} & \textbf{0.5611} \\ \hline
	\end{tabular}}
		\vspace*{-0.1in}
	
	\label{table:WFLW}
\end{table*}

\begin{table}
	%	\vspace*{-0.15in}
	\centering
	\setlength\tabcolsep{2.5pt}
	\caption{Evaluation on the 300W dataset.
		Key=[\textcolor{red}{\textbf{Best}}, \textcolor{blue}{\textbf{Second Best}}]. * indicates the SCBE module is trained on 300W dataset.}
	\resizebox{\linewidth}{!}{%
		\begin{tabular}{c|c|c|c}
			\hline
			Method  & 
			\begin{tabular}[c]{@{}c@{}}Common\\ Subset\end{tabular} & \begin{tabular}[c]{@{}c@{}}Challenging\\ Subset\end{tabular} & Fullset \\ \hline
			\multicolumn{4}{c}{Inter-pupil Normalization}                                                                                             \\ \hline 
			CFAN\textsubscript{ECCV 14}~\cite{zhang2014coarse}        & 5.50    & 16.78     & 7.69    \\
			SDM\textsubscript{CVPR 13}~\cite{xiong2013supervised}     & 5.57    & 15.40     & 7.50    \\
			LBF\textsubscript{CVPR 14}~\cite{ren2014face}             & 4.95    & 11.98     & 6.32    \\
			CFSS\textsubscript{CVPR 15}~\cite{zhu2015face}            & 4.73    & 9.98      & 5.76    \\
			TCDCN\textsubscript{16'}~\cite{zhang2015learning}         & 4.80    & 8.60      & 5.54    \\
			MDM\textsubscript{CVPR 16}~\cite{trigeorgis2016mnemonic}  & 4.83    & 10.14     & 5.88    \\
			RAR\textsubscript{ECCV 16}~\cite{xiao2016robust}          & 4.12    & 8.35      & 4.94    \\
			DVLN\textsubscript{CVPR 17}~\cite{wu2017leveraging}       & 3.94    & 7.62      & 4.66    \\
			TSR\textsubscript{CVPR 17}~\cite{lv2017deep}              & 4.36    & 7.56      & 4.99    \\
			DSRN\textsubscript{CVPR 18}~\cite{miao2018direct}         & 4.12    & 9.68      & 5.21    \\
			\small{RCN\textsuperscript{+}(L+ELT)\textsubscript{CVPR 18}}~\cite{honari2018improving}  & 4.20 & 7.78  & 4.90  \\
			DCFE\textsubscript{ECCV 18}~\cite{valle2018deeply}        & 3.83    & 7.54      & 4.55 \\
			LAB\textsubscript{CVPR 18}~\cite{wayne2018lab}            & \textcolor{blue}{\textbf{3.42}}       &  6.98       & 4.12    \\
			Wing\textsubscript{CVPR 18}~\cite{feng2018wing}           & \textcolor{red}{\textbf{3.27}}        & 7.18        & \textcolor{red}{\textbf{4.04}} \\
			AWing \textsubscript{ICCV 19}~\cite{wang2019adaptive}     & 3.77    & \textcolor{red}{\textbf{6.52}}  & \textcolor{blue}{\textbf{4.31}}  \\\hline
			\textbf{TAB(Ours)}                                        & 3.81    & \textcolor{blue}{\textbf{6.76}} & 4.39    \\ \hline
			\multicolumn{4}{c}{Inter-ocular Normalization}                                                       \\ \hline
			PCD-CNN\textsubscript{CVPR 18}~\cite{kumar2018disentangling} & 3.67 & 7.62      & 4.44    \\
			SAN\textsubscript{CVPR 18}~\cite{dong2018style}           & 3.34    & 6.60      & 3.98    \\
			LAB\textsubscript{CVPR 18}~\cite{wayne2018lab}            & 2.98    & 5.19      & 3.49    \\
			DU-Net\textsubscript{ECCV 18}~\cite{tang2018quantized}    & 2.90    & 5.15      & 3.35    \\
			HRNet\textsubscript{19'}~\cite{sun2019high}               & 2.87    & 5.15      & 3.32    \\
			AWing \textsubscript{ICCV 19}~\cite{wang2019adaptive}     & \textcolor{red}{\textbf{2.72}}  & \textcolor{red}{\textbf{4.52}}   & \textcolor{red}{\textbf{3.07}} \\
			LUVLi \textsubscript{CVPR 20}~\cite{kumar2020luvli}       & {2.76}  & {5.16}    & {3.23}  \\ \hline
			\textbf{TAB$^*$(Ours)}                                        & \textcolor{blue}{\textbf{2.75}} & 4.74  & 3.14 \\
			\textbf{TAB(Ours)}                                        & \textcolor{blue}{\textbf{2.75}} & \textcolor{blue}{\textbf{4.68}}  & \textcolor{blue}{\textbf{3.13}} \\\hline
	\end{tabular}}
	
	\label{table:300W}
	%	\vspace*{-0.15in}
\end{table}
With multi-level boundary information, our method can obtain more uniformed landmark predictions than other methods, as shown in Fig.~\ref{fig:transition_example}. Take the face in the third row for example, the shape of landmarks predicted by our approach is more smoother and fit better the chin than that predicted by HRNet~\cite{sun2019high} and AWing~\cite{wang2019adaptive}. While landmark drifting and big interval (red dashed circle) occurs for HRNet, due to the occlusion of mask, our approach is more robust against such occlusion. 

\section{Experiments}

\begin{table}
	\centering
%	\footnotesize
\setlength\tabcolsep{2.5pt}
	\caption{Evaluation on 300W private dataset. Key:[\textcolor{red}{\textbf{Best}}, \textcolor{blue}{\textbf{Second Best}}]}
	\resizebox{\linewidth}{!}{%
	\begin{tabular}{c|c|c|c} %表格7列 全部居中显示
		\hline
		Method &NME(\%)($\downarrow$) & AUC\textsubscript{8\%}(\%)($\uparrow$) & FR\textsubscript{8\%}(\%)($\downarrow$)  \\ \hline
		ESR\textsubscript{CVPR 14}~\cite{cao2014face}                 & -         & 32.35     & 17.00     \\
		cGPRT\textsubscript{CVPR 15}~\cite{lee2015face}               & -         & 41.32     & 12.83     \\
		CFSS\textsubscript{CVPR 15}~\cite{zhu2015face}             & -         & 39.81     & 12.30     \\
		MDM\textsubscript{CVPR 16}~\cite{trigeorgis2016mnemonic}                 & 5.05      & 45.32     & 6.80      \\
		DAN\textsubscript{CVPRW 17}~\cite{kowalski2017deep}                 & 4.30      & 47.00     & 2.67      \\
		SHN\textsubscript{CVPRW 17}~\cite{yang2017stacked}                 & 4.05      & -         & -         \\
		DCFE\textsubscript{ECCV 18}~\cite{valle2018deeply}                & 3.88      & 52.42     & 1.83      \\
		AWing\textsubscript{ICCV 19}~\cite{wang2019adaptive}      & \textcolor{red}{\textbf{3.56}} & \textcolor{red}{\textbf{55.76}} & \textcolor{blue}{\textbf{0.83}} \\
		\hline
		\textbf{TAB(Ours)}      &\textcolor{blue}{\textbf{3.59}} & \textcolor{blue}{\textbf{55.20}} & \textcolor{red}{\textbf{0.50}} \\ \hline
		& NME(\%)($\downarrow$)       & AUC\textsubscript{10\%}(\%)($\uparrow$)       & FR\textsubscript{10\%}(\%)($\downarrow$)        \\ \hline
		M3-CSR\textsubscript{16'}~\cite{deng2016m3}         & -          & 47.52    & 5.5       \\
		Fan \etal \textsubscript{16'}~\cite{fan2016approaching} & -          & 48.02    & 14.83     \\
		DR + MDM \textsubscript{CVPR 17}~\cite{alp2017densereg}      & -          & 52.19    & 3.67      \\
		JMFA\textsubscript{17'}~\cite{deng2019joint}                & -          & 54.85    & 1.00      \\
		LAB\textsubscript{CVPR 18}~\cite{wayne2018lab}                 & -          & 58.85    & 0.83      \\
		AWing\textsubscript{ICCV 19}~\cite{wang2019adaptive}      & \textcolor{red}{\textbf{3.56}} & \textcolor{red}{\textbf{64.40}} & \textcolor{blue}{\textbf{0.33}} \\ \hline
		\textbf{TAB(Ours)}      & \textcolor{blue}{\textbf{3.59}} & \textcolor{blue}{\textbf{64.11}} & \textcolor{red}{\textbf{0.17}} \\ \hline		
	\end{tabular}}

	\vspace*{-0.1in}
	\label{table:300WPrivate}
\end{table}

\begin{table}
	\centering
	\setlength\tabcolsep{3pt}
	\caption{Evaluation on the COFW dataset.}
	\resizebox{\linewidth}{!}{%
		\begin{tabular}{c|c|c|c}
			\hline
			Method              & NME       & AUC\textsubscript{10\%}($\uparrow$)       & FR\textsubscript{10\%}($\downarrow$)        \\ \hline
			Human~\cite{burgos2013robust}                 & 5.60         & -     & 0.00     \\
			Wu \etal\textsubscript{ICCV 15}~\cite{wu2015robust}        & 5.93      & -     & -      \\
			RAR\textsubscript{ECCV 16}~\cite{xiao2016robust}                & 6.03         & -     & 4.14     \\
			DAC-CSR\textsubscript{CVPR 17}~\cite{feng2017dynamic}       & 6.03      & -     & 4.73      \\
			SHN\textsubscript{CVPRW 17}~\cite{yang2017stacked}         & 5.60      & -         & -         \\ 
			PCD-CNN\textsubscript{CVPR 18}~\cite{kumar2018disentangling}   & 5.77      & -         & 3.73         \\
			Wing\textsubscript{CVPR 18}~\cite{feng2018wing}  & 5.44 & - &	3.75 \\ 
			{AWing}\textsubscript{ICCV 19}~\cite{wang2019adaptive}        & {4.94} & {48.82} & {0.99} \\\hline \textbf{TAB(Ours)}      & \textbf{4.39} & \textbf{56.14} & \textbf{0.20} \\ \hline
			& NME       & AUC\textsubscript{8\%}($\uparrow$)       & FR\textsubscript{8\%}($\downarrow$)       \\ \hline
			DCFE\textsubscript{ECCV 18}~\cite{valle2018deeply}     & 5.27          & 35.86    & 7.29      \\ 
			{AWing}\textsubscript{ICCV 19}~\cite{wang2019adaptive}      & {4.94} & {39.11} & {5.52} \\\hline
			\textbf{TAB(Ours)}      & \textbf{4.39} & \textbf{45.29} & \textbf{0.79} \\ \hline
	\end{tabular}}
	
%	\vspace*{-0.15in}
	\label{table:cofw}
\end{table}

\begin{table}
	
	\centering
	\footnotesize
	\caption{Evaluation on COFW-68 dataset.
		[Key:*=Pretrained on 300W-LP-2D]}
	\resizebox{\linewidth}{!}{%
	\begin{tabular}{c|c|c} %表格7列 全部居中显示
		\hline
		Method &NME(\%)($\downarrow$)& AUC\textsubscript{10\%}(\%)($\uparrow$) \\
		\hline
		SAN\textsuperscript{*}\textsubscript{CVPR 18}~\cite{dong2018style} & 3.50 & 51.9 \\
		2D-FAN\textsuperscript{*}\textsubscript{ICCV 17}~\cite{bulat2017far} & 2.95 & 57.5 \\
		Softlabel\textsuperscript{*}\textsubscript{ICCV 19}~\cite{chen2019face} & 2.92 & 57.9 \\
		KDN\textsuperscript{*}\textsubscript{ICCV 19}~\cite{chen2019face} & 2.73 & 60.1 \\
		LUVLi\textsuperscript{*}\textsubscript{CVPR 20}~\cite{kumar2020luvli} & 2.57 & 63.4 \\ \hline  %纵向合并4行单元格 
		%为第二列到第七列添加横线
		\textbf{TAB(Ours)} & \textbf{2.43} & \textbf{65.4}\\
		
		\hline
	\end{tabular}}
	
	\label{table:cofw68}
\end{table}
\begin{figure}
	\begin{center}
		%\fbox{\rule{0pt}{2in} \rule{.9\linewidth}{0pt}}
		\includegraphics[width=\linewidth]{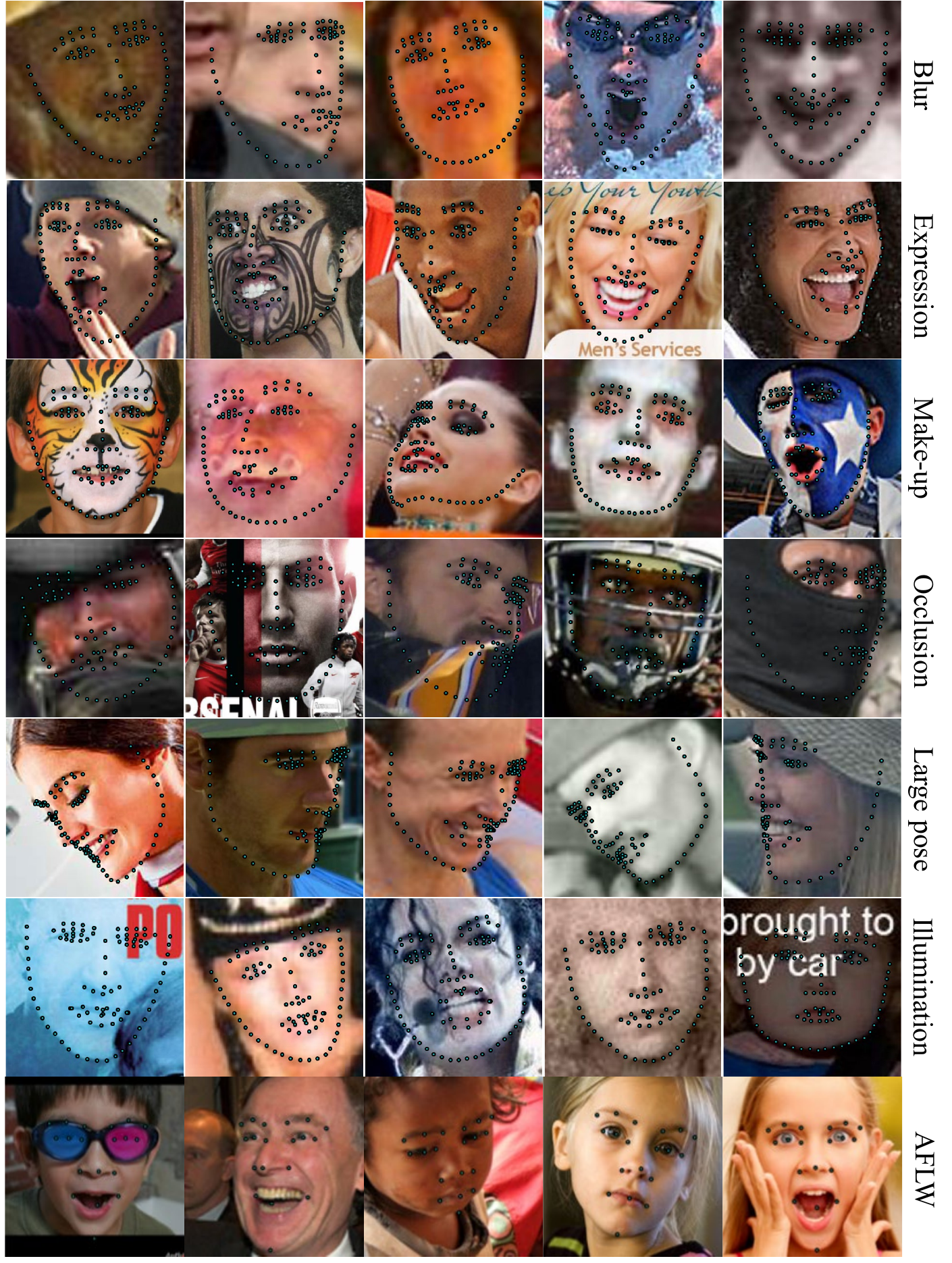}
	\end{center}
	%	\vspace*{-0.2in}
	\caption{Example results on WFLW six test subsets and AFLW dataset,  which justify the robustness of our method to faces with blur, expression, make-up, heavy occlusion, illumination, and large poses.}
	\label{fig:short}
	%	\vspace*{-0.2in}
\end{figure}
 \begin{table}
% \vspace*{-0.1in} 
 \centering
 \caption{Evaluation on AFLW dataset.}
 \resizebox{\linewidth}{!}{%
 \setlength\tabcolsep{1.5pt}
 
 \begin{tabular}{c|c|c}
 \hline
 Method  & Full(\%)($\downarrow$) & Frontal(\%)($\downarrow$) \\ \hline
 CFSS\textsubscript{CVPR 15}~\cite{zhu2015face}    &    3.92       &     2.68         \\
 CCL\textsubscript{CVPR 16}~\cite{zhu2016unconstrained}     &    2.72       &     2.17         \\
 TSR\textsubscript{CVPR 17}~\cite{lv2017deep}     &    2.17       &      -           \\
 DAC-OSR\textsubscript{CVPR 17}~\cite{feng2017dynamic} &    2.27       &     1.81         \\
 LLL\textsubscript{ICCV 19}~\cite{robinson2019laplace}  &       1.97 & - \\
 SAN\textsubscript{CVPR 18}~\cite{dong2018style} & 1.91 & 1.85 \\
 DSRN\textsubscript{CVPR 18}~\cite{miao2018direct} & 1.86 & - \\
 LAB\textsubscript{CVPR 18}~\cite{wayne2018lab}     &    1.85       &     1.62         \\
 HRNet\textsubscript{19'}~\cite{sun2019high}    &    1.57       &      1.46           \\
 LUVLi\textsubscript{CVPR 20}~\cite{kumar2020luvli}    &    \textbf{1.39}       & \textbf{1.19}                 \\\hline
 \textbf{TAB(Ours)} & 1.47 & 1.25 \\ \hline
 \end{tabular}}
 
 \label{table:AFLW}
% \vspace*{-0.15in}
 \end{table}

\begin{table}
%	\vspace*{0.15in}
	\caption{Evaluation on AFLW2000-3D dataset(68 2D landmarks). The NME for faces with different yaw angles are reported.}
		\resizebox{0.48 \textwidth}{!}{
			\begin{tabular}{c|c|c|c|c}
				\hline
				\multirow{2}*{Methods} & \multicolumn{4}{c}{\text{NME(\%)($\downarrow$)}} \\
				\cline{2-5}
				&$0^{\circ}$ to $30^{\circ}$&$30^{\circ}$ to $60^{\circ}$ &$60^{\circ}$ to $90^{\circ}$&Mean \\
				\hline
				SDM\textsubscript{CVPR 13}~\cite{xiong2013supervised}& 3.67 & 4.94 & 9.67 & 6.12 \\
				3DDFA\textsubscript{CVPR 16}~\cite{zhu2016face3d} & 3.78 & 4.54 & 7.93 & 5.42 \\
				3DDFA + SDM\textsubscript{CVPR 16}~\cite{zhu2016face3d} & 3.43 & 4.24 & 7.17 & 5.42 \\
				Yu \textit{et al.}\textsubscript{ICCV 17}~\cite{yu2017learning} & 3.62 & 6.06 & 9.56 & - \\
				3DSTN\textsubscript{ICCV 17}~\cite{bhagavatula2017faster} & 3.15 & 4.33 & 5.98 & 4.49 \\
				DeFA\textsubscript{ICCVW 17}~\cite{liu2017dense} & - & - & - & 4.50 \\
				PRNnet\textsubscript{ECCV 18}~\cite{feng2018joint}& 2.75 & 3.51 & 4.61 & 3.62 \\
				2DASL\textsubscript{TMM 20}~\cite{tu20203d}& 2.75 & 3.44 & \textbf{4.41} & 3.53\\
				\hline
				\textbf{TAB(Ours)}& \textbf{2.44} & \textbf{3.23} & 4.60 & \textbf{3.42}\\
				\hline
		\end{tabular}}
%	\vspace{0.05in}

	\label{table:aflw2000-3d}
%	\vspace{-0.1in}
\end{table}

\subsection{Dataset} We train and evaluate our method on the following popular and challenging face alignment datasets including \textbf{WFLW}~\cite{wayne2018lab},  \textbf{300W}~\cite{sagonas2013300}, 
\textbf{300W-LP} and \textbf{AFLW2000-3D}~\cite{zhu2016face3d}, \textbf{AFLW}~\cite{koestinger2011annotated}, \textbf{COFW}~\cite{ghiasi2015occlusion} and \textbf{COFW-68}~\cite{ghiasi2015occlusion}.

%WFLW~\cite{wayne2018lab}, 300W~\cite{sagonas2013300}, COFW~\cite{burgos2013robust}, COFW-68~\cite{ghiasi2015occlusion}, AFLW~\cite{koestinger2011annotated} and 3D face alignment benchmark AFLW2000-3D~\cite{zhu2016face3d}.

\textbf{WFLW dataset.} It is a challenging 98 facial landmarks localization dataset based on the widerface dataset~\cite{yang2016wider}. There are 10000 faces, we use 7500 faces for training and 2500 faces for testing. The test set is split into six subsets for evaluation which presents variations of large pose (326 images), expression (314 images),  illumination(698 images), blur(773 images), make-up(206 images) and occlusion(736 images).

\textbf{300W dataset.} It is a widely used benchmark for evaluating facial landmarks localization algorithms. There are 3148 face images for training. The test set contains 689 faces which are split into a common subset (554 images) and a challenging subset (135 images) for validation. The test set for the competition of 300W contains 600 image samples. Each face is annotated with 68 landmarks.

\textbf{300W-LP} and \textbf{AFLW2000-3D}. 300W-LP is a synthetic dataset based on 300W dataset annotated with 68 landmarks and it contains 61225 face images with large poses ranging from $0^\circ$ to $90^\circ$. AFLW2000-3D contains 2000 face images with 68 annotated landmarks. In our experiments, 300W-LP is used for training and AFLW2000-3D dataset is used for testing.

\textbf{COFW} and \textbf{COFW-68}. COFW contains 1345 images for training and 507 images for testing. Each face is annotated with 29 landmarks. COFW-68 is a reannotated dataset of COFW using the 68-landmarks annotation format of the 300W dataset.  

\textbf{AFLW}. The widely used dataset consists of 20000 images for training. Two test sets, i.e. aflw-full(4386 images) and frontal (1314 images selected from aflw-full) are available for evaluation. Each face is annotated with 19 landmarks.

\subsection{Evaluation metrics}

\textbf{Normalized Mean Error (NME)} is a point-to-point normalized euclidian distance. NME for each image is calculated by the following formula:

\begin{equation}
NME(P, \hat{P}) = \frac{1}{L}\sum_{i=1}^{L}\frac{\vert\vert p_i - \hat{p}_i\vert\vert}{d}
\end{equation}
where $P$ and $\hat{P}$ are the ground truth and predicted coordinates of landmarks. $L$ is the number of landmarks. $p_i$ and $\hat{p}_i$ are the ground-truth and predicted coordinates of the $ith$ landmark. $d$ is the normalization factor. For the WFLW dataset, we follow~\cite{wayne2018lab} to use the \textit{inter-ocular}(outer-eye-corner distance) as the normalization factor.  \textit{Inter-pupil}(eye-center distance) is used for the COFW and 300W datasets. We use the face size $\sqrt{h_{bbox}*w_{bbox}}$ as the factor $d$ for AFLW, AFLW2000-3D and COFW-68 dataset.

\textbf{Falure Rate (FR).} If an image's NME is larger than a threshold, it is considered as a failure case. For the WFLW dataset, we follow~\cite{wayne2018lab} and choose 10\% as the threshold. For 300W, we follow~\cite{wang2019adaptive} and choose 8\% and 10\% as the thresholds, respectively. 

\textbf{Cumulative Error Distribution (CED).} We follow~\cite{wayne2018lab, wang2019adaptive, sun2019high} to plot the the CED cure. And Area Under Curve (AUC) is calculated based on the CED curve. Larger AUC means that a larger portion of the test dataset is well predicted.

\subsection{Implementaion details}
We cropped each image by the bounding box of face provided by HRNet~\cite{sun2019high} and resized it to 256$\times$256 as input. The SCBE module consists of 2-Stacked Hourglass Network~\cite{bulat2017far} with vgg16bn stem layers(implemented by Torchvision) and the BALT module was based on UNet. Data augmentation was performed with flipping(50\%), random rotation($\pm 60^\circ$), rescaling($\pm$25\%), random coarse dropout(50\%) and random brightness or darkness. During training, we chose Adam optimizer with initial learning rate 0.0001, and set the weight decay to 0. We  trained the SCBE module for 140 epoch with multiplication of 0.1 at 80 and 120 epoch.
Then we froze the paramters of the SCBE module to train the whole model with the same setting. We used a normal distribution with $ std=0.001$ and $mean=0$ to initialize the parameters of stacked hourglass. The parameters of vgg16bn pre-trained on ImageNet were used to initialize stem layers. Mean quared error was adopted as the loss function to train these two modules. All experiments were conducted on  PyTorch~\cite{paszke2019pytorch}  with one Tesla P100 GPU(16GB). While the SCBE modules for WFLW and 300W datasets are trained using their corresponding training set, that for other datasets are directly taken from the one trained by WFLW dataset. 

\begin{figure}
	\begin{center}
		%\fbox{\rule{0pt}{2in} \rule{.9\linewidth}{0pt}}
		\includegraphics[width=\linewidth]{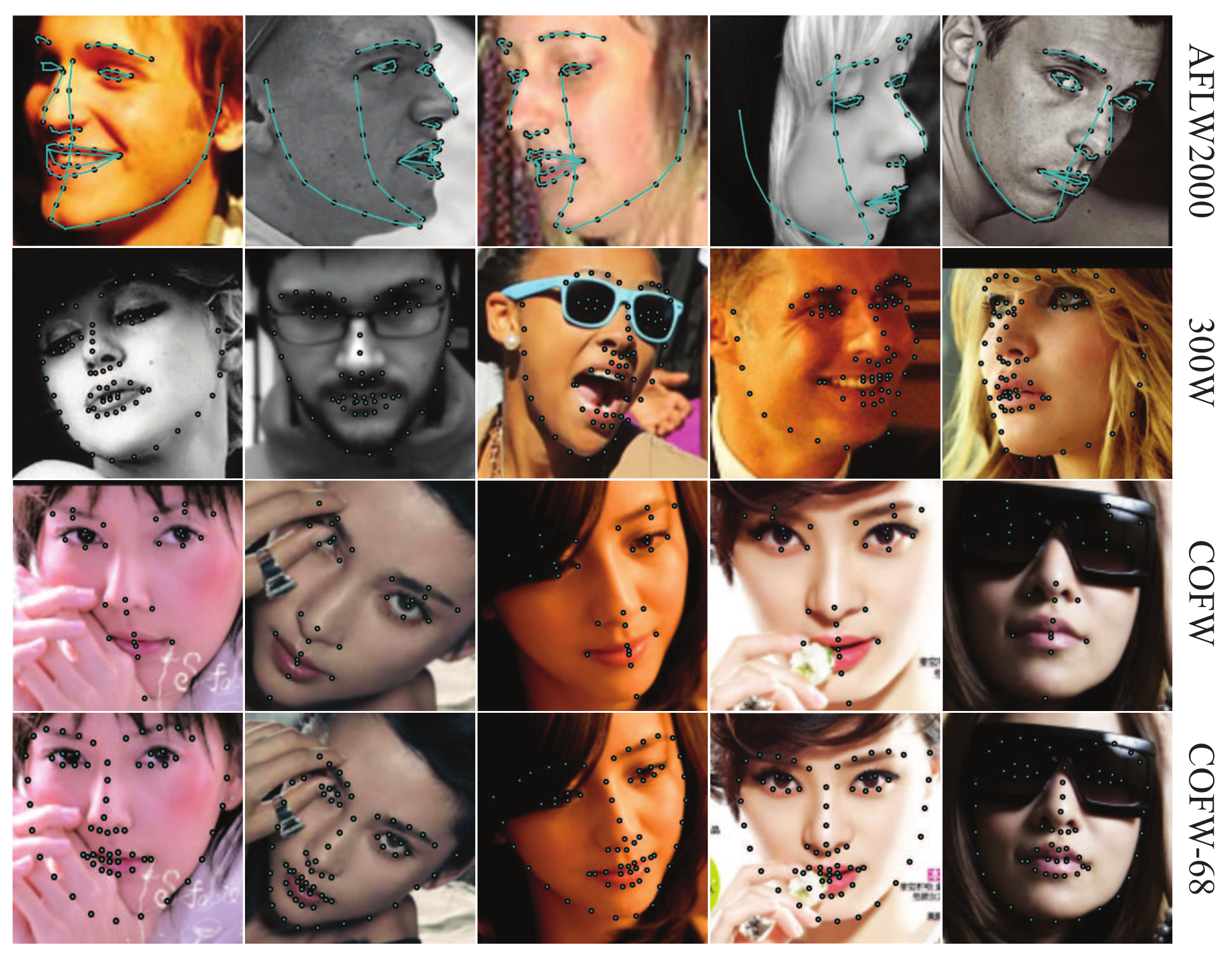}
	\end{center}
	%	\vspace*{-0.2in}
	\caption{Example results on AFLW2000, 300W, COFW, and COFW-68 datasets.}
	\label{fig:examplar2}
	%	\vspace*{-0.2in}
\end{figure}
\begin{figure}
	\begin{center}
		%\fbox{\rule{0pt}{2in} \rule{.9\linewidth}{0pt}}
		\includegraphics[width=\linewidth]{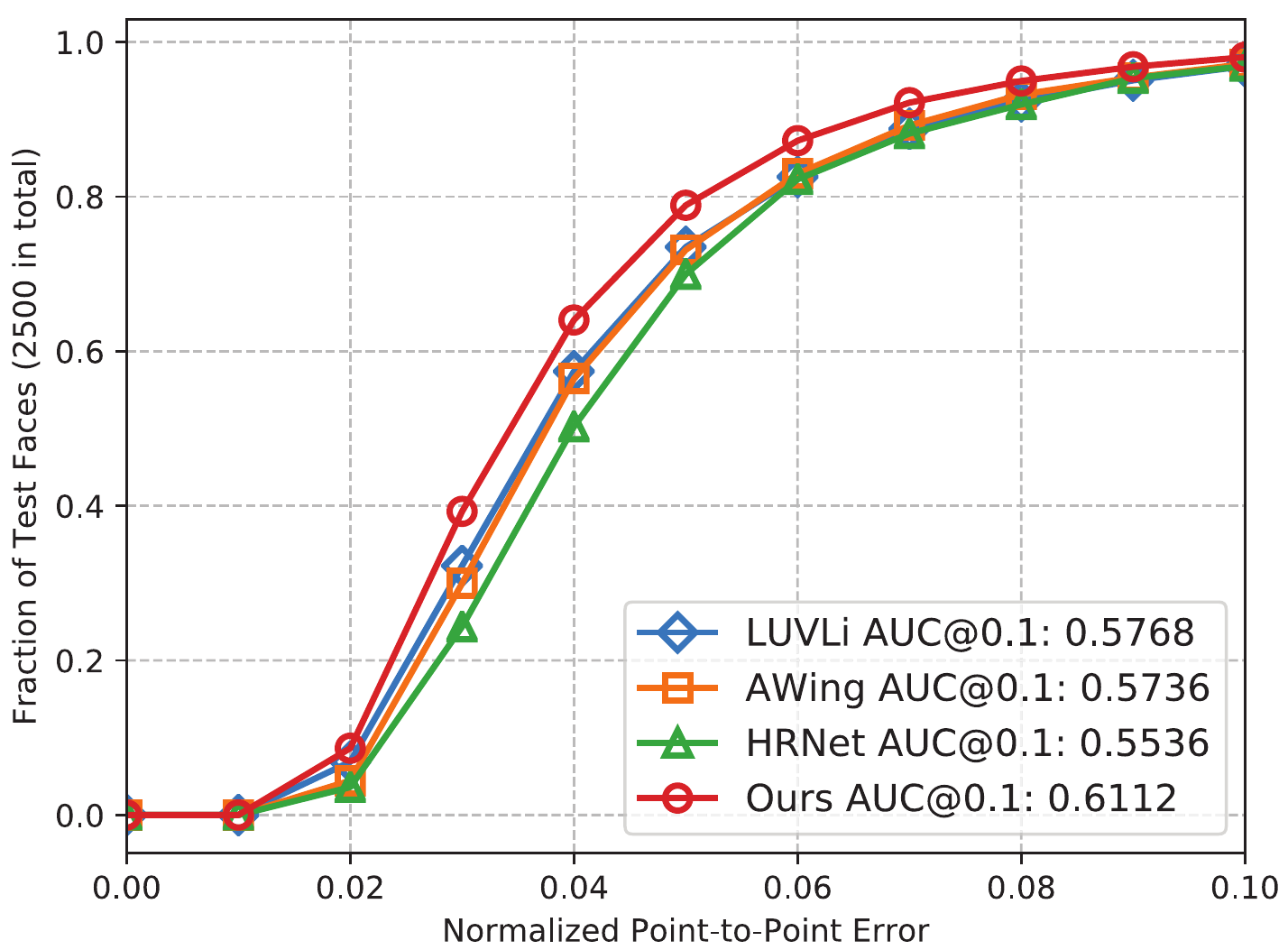}
	\end{center}
	%	\vspace*{-0.2in}
	\caption{CED for WFLW testset (98 landmarks). AUC@0.1 is also reported.}
	\label{fig:ced}
		\vspace*{-0.1in}
\end{figure}
\subsection{WFLW dataset}
The results on the WFLW dataset are shown in Table~\ref{table:WFLW} and Fig~\ref{fig:ced}. Compared to state-of-the-art methods such as HRNet, AWing, and LUVLi, our approach displays a significant improvement on the WFLW test set and the other six test subsets, in terms of Normalized Mean Error, Failure Rate, and Area Under the Curve. These evaluation results verify that our SCBE module and BALT module can accurately extract boundary information and efficiently transform it to landmark heatmaps. It also indicates the robustness of our method for face alignment in the wild.

\subsection{300W dataset} 
The results on 300W dataset and private set are shown in Table~\ref{table:300W} and Table~\ref{table:300WPrivate}. To evaluate the capacity of our SCBE module, we also train it on the WFLW dataset and apply it to the 300W dataset for face alignment. Our method directly achieves comparative result and ranks the second-best in terms of the inter-ocular metric, which verifies the effiveness of our SCBE module and BATL module.

\subsection{COFW and COFW-68 datasets}
On the COFW dataset, the network is trained on the COFW training set and tested on the COFW test set. On the COFW-68 dataset, the network is trained on the 300W training set and tested on the COFW-68 test set. The results on COFW and COFW-68 datasets are shown in Table~\ref{table:cofw}, Table~\ref{table:cofw68}, and Fig ~\ref{fig:examplar2}. As shown in the results, our method outperforms all state-of-the-art methods by a significant margin. It verifies that our SCBE module and BALT module have a good generalization on faces with different number of facial landmarks.

\subsection{AFLW dataset}
Table~\ref{table:AFLW} provides the results on AFLW dataset. Our method achieves the second-best performance among state-of-the-art methods. As only 19 landmarks are annotated for this dataset, there might be large intervals between landmarks. The large gap may interfere with the shape of facial boundary and compromise the performance of following landmark heatmap regression.

\subsection{AFLW2000-3D dataset}
Table~\ref{table:aflw2000-3d} provides the results on AFLW2000-3D test set. For the faces with yaw angle from $0^{\circ}$ to $60^{\circ}$, our method achieves the best performance among all avalible models. Although the NME of our approach for faces with yaw angle from $60^{\circ}$ to $90^{\circ}$ is a little bit bigger than that of 2DASL~\cite{tu20203d}, the mean NME of our approach still ranks the top. The results clearly suggest that our approach is robust against the large pose variations.

\subsection{Ablation study}\label{subsection:as}

In this section, we use 300W challenging subset to validate the effectiveness of the stem layers and multi-scale features fusion, and evaluate different combinations of feature selection, i.e. single-scale (ss) and multi-scale (ms), and feature fusion, i.e. $0\times, 1\times, 2\times, 3\times$. 

\textbf{Stem layers.} Robust stem layers extract more rich and accurate local features, which boost the performance of the SCBE module. We tested different stem layers like ResNet-50 and VGG, and compared their performances in Table~\ref{table:stemlayers} with the baseline, which adopt the stem layers of the Hourglass networks. As shown in Table~\ref{table:stemlayers}, the MSE, AUC, NME and FR of VGG achieved better performance than ResNet and baseline. 

\begin{table}
	\centering
	\footnotesize
	\caption{Evaluation on 300W challenging subset with different stem layers. MSE indicates the mean squared error between ground-truth and predicted boundary heatmaps.}
	\begin{tabular}{c|c|c|c|c} %表格7列 全部居中显示
		\hline
		Stem & MSE & AUC\textsubscript{10\%}(\%)($\uparrow$) &NME(\%)($\downarrow$) & FR\textsubscript{8\%}(\%)($\downarrow$)  \\
		\hline
		baseline & 0.002536 & 51.54 & 4.86 & 2.96 \\  %纵向合并4行单元格 
		%为第二列到第七列添加横线
		resnet & 0.002503 & 52.36 & 4.77 & 2.22 \\
		
		\textbf{vgg} & \textbf{0.002442} & \textbf{52.98} & \textbf{4.70} & \textbf{2.22} \\
		
		\hline
		%		\multicolumn{2}{c|}{no} & vgg & 0.002442 & & \\
		%		\hline
	\end{tabular}
	
	\label{table:stemlayers}
%	\vspace*{-0.1in}
\end{table}

\begin{table}
	\centering
	\footnotesize
	\caption{Evaluation on 300W challenging subset with different feature selections.}
	\resizebox{\linewidth}{!}{%
	\begin{tabular}{c|c|c|c} %表格7列 全部居中显示
		\hline
		Feature & AUC\textsubscript{10\%}(\%)($\uparrow$) &NME(\%)($\downarrow$) & FR\textsubscript{8\%}(\%)($\downarrow$)  \\
		\hline
		ss & 52.24 & 4.78 & 2.22 \\  %纵向合并4行单元格 
		%为第二列到第七列添加横线
		\textbf{ms} & \textbf{52.98} & \textbf{4.70} & \textbf{2.22} \\
		
		\hline
	\end{tabular}}
	
	\label{table:feature_selection}
%	\vspace*{-0.1in}
\end{table}
\begin{table}
	\centering
	\footnotesize
	\caption{Evaluation on 300W challenging subset with different feature fusion settings.}
	\resizebox{\linewidth}{!}{%
	\begin{tabular}{c|c|c|c} %表格7列 全部居中显示
		\hline
		Fusion & AUC\textsubscript{10\%}(\%)($\uparrow$) &NME(\%)($\downarrow$) & FR\textsubscript{8\%}(\%)($\downarrow$)  \\
		\hline $0\times$ &51.35 & 4.87 &  3.70\\ 
		 $1\times$ & 52.56 & 4.74 & 2.22\\
		 \textbf{$2\times$} & \textbf{52.98} & \textbf{4.70} & \textbf{2.22} \\
		
		$3\times$ & 52.68 & 4.73 & 2.22 \\
		\hline
	\end{tabular}}
	\label{table:fusionmanner}
%	\vspace*{-0.1in}
\end{table}

\textbf{Boundary-aware features selection.} Using features extracted by the SCBE module as additional information not only accelerates the convergence of the BALT module, but also enhances the shape constraint of facial boundary on landmarks. How to choose these features? We have tried single-scale feature (ss) and multi-scale features (ms) selected from the last hourglass network. While single-scale feature only uses the output of the last convolutional layers before heatmap predictions, multi-scale features fuse the s different output of the last hourglass network. We report the AUC, NME, and FR of these selections in Table~\ref{table:feature_selection}. It can be observed that multi-scale features are better than single-scale feature.

\textbf{Features fusion.} We tried $0\times, 1\times, 2\times, \text{and}\, 3\times$ features fusion settings in the BALT module (note that, 0$\times$ means no features fusion in the encoding process). We report the AUC, NME, and FR of these settings in Table~\ref{table:fusionmanner}. It can be observed that the model with features fusion $1\times, 2\times, \text{and}\, 3\times$ generally achieves better performance than that without feature fusion ($0\times$). While same FR was achieved for different fusion settings, $2\times$ strategy achieved the best AUC and NME.

\section{Discussion and Conclusion}
%-------------------------------------------------------------------------

In this paper, we present a two-stage but end-to-end method to deal with issues of occlusion, large pose, and expression, etc., in face alignment task. By leveraging the geometric structure of faces and the spatial information of heatmaps, our method can easily learn a shape constrained transformation to get boundary-aware landmark heatmap predictions. Experiment results on widely used datasets verify the effectiveness and great potential of our approach. When GeForce GTX TITAN X GPU(12GB) is concerned, the size and processing speed of our model with 2-stacked hourglass are 86.9MB and 27ms, respectively; the size and processing speed for 1-stacked hourglass are 64.5MB and 17ms, respectively. 

We believe the unique facial structure is the key to localize facial landmarks~\cite{wayne2018lab}, due to the fact that (1) The number of annotated facial landmarks is increasing and getting denser, which fits the shape of face better. (2) The facial structure information is more likely to be reserved and estimated on unconstrained scenes. Considerable attention could be paid to the analysis between facial structure and facial landmarks for further improvements in face alignment task.

{\small
\bibliographystyle{ieee_fullname}
\bibliography{egbib}
}
\end{document}